\documentclass[11pt,a4paper,twocolumn]{article}

\usepackage[utf8]{inputenc}
\usepackage[T1]{fontenc}

\usepackage[a4paper, textwidth=43pc, vmargin=25mm]{geometry}
\usepackage{setspace}
\usepackage{amsmath}
\usepackage{amssymb}
\usepackage{amsfonts}

\usepackage{graphicx}
\usepackage{subcaption}
\usepackage{placeins}

\usepackage{booktabs}
\usepackage{multirow}
\usepackage{makecell}
\usepackage{array}

\usepackage{textcomp}
\usepackage{url}

\usepackage[numbers,sort&compress]{natbib}

\usepackage{xcolor}
\usepackage[hidelinks]{hyperref}

\title{Illumination-Robust Camera-Based Heart-Rate Estimation for Physiological Sensing in Robots}

\author{%
  Zhi Wei Xu\thanks{%
    Department of Mechanical Engineering, National Cheng Kung University, Tainan, Taiwan.
    ORCID: \href{https://orcid.org/0009-0007-1471-7554}{0009-0007-1471-7554}.
    Email: \texttt{andy.xu@nordlinglab.org}.%
  }%
  \and
  Torbj{\"o}rn E.\,M.\ Nordling\thanks{%
    Department of Mechanical Engineering, National Cheng Kung University, Tainan, Taiwan.
    ORCID: \href{https://orcid.org/0000-0003-4867-6707}{0000-0003-4867-6707}.
    Corresponding author: \texttt{torbjorn.nordling@nordlinglab.org}.%
  }%
}

\date{\today}

\begin{document}
\maketitle

\begin{abstract}
Physiological awareness is important for service, social, and assistive robots that interact with humans in everyday environments.
Remote photoplethysmography (rPPG) enables non-contact heart-rate (HR) estimation from an RGB camera, making it a promising sensing modality for robot-mounted vision systems.
However, illumination variation remains a major barrier to robust deployment.
This paper presents an end-to-end spatial-temporal transformer framework for remote HR estimation on a new dataset with varied illumination.
Our estimator integrates PRNet-based 3D face alignment, clip-level illumination augmentation, the Residual Temporal Standardization Module, and controlled hybrid temporal-frequency supervision.
The training objective combines a Soft-Shifted Pearson waveform loss with a spectral Kullback-Leibler divergence loss, where a tuned weight ($\mathbf{\beta}$) controls the contribution of frequency-domain heart-rate guidance.
Experiments on a static all-level mix protocol covering three illumination levels show that $\mathbf{\beta=5}$ provides the strongest result among the tested beta settings, achieving a best-run HR mean absolute error (MAE) of 0.79 bpm and an HR correlation of 0.982.
Compared with the PhysFormer baseline evaluated on our dataset, our estimator reduces HR MAE by 93.6\%, while increasing HR correlation from 0.088 to 0.982, making it usable when illumination varies.

\end{abstract}


\bigskip

\section{Introduction}
Robots that interact with humans in service, healthcare, and assistive scenarios need to estimate not only external behavior but also internal physiological state.
Heart rate (HR) is a useful physiological cue for understanding human condition, workload, stress, and safety, but wearable or contact sensors would require data sharing with robots, which is unlikely.
Robots already have RGB cameras.
Camera-based physiological sensing therefore provides an attractive path toward unobtrusive human-aware robot perception \cite{savur2023physiological}.

Remote photoplethysmography (rPPG) aims to recover pulse-related physiological signals from ordinary facial videos by analyzing subtle temporal color variations caused by blood volume changes beneath the skin\cite{verkruysse2008remote}.
Because rPPG only requires an RGB camera, it is well aligned with human-facing robots that already use cameras for perception and interaction \cite{stricker2014mobile}.
However, the useful rPPG component is weak compared with environmental and acquisition noise.
In real human-robot interaction environments, camera observations may be captured under different room lighting, viewing angles, and user positions.
Cross-video illumination variation can dominate the weak pulse-related color signal, making robust camera-based physiological perception difficult.

Heart-rate information appears in both the temporal waveform and the spectral peak of the rPPG signal.
Prior rPPG studies therefore increasingly combine temporal and frequency-domain supervision.
AutoHR combines Pearson-style temporal waveform constraints with frequency-domain HR supervision \cite{yu2020autohr}.
PhysFormer integrates temporal waveform supervision, frequency cross-entropy, and label-distribution learning, and further studies frequency weighting and dynamic weighting effects \cite{yu2022physformer}.
LSTC-rPPG uses time-domain waveform regression with a scale-invariant power spectral density (PSD) loss \cite{lee2024lstcrppg}, while recent robust rPPG frameworks use temporal, frequency, and HR-distribution supervision or dynamic hybrid losses to improve generalization in complex and cross-dataset scenarios \cite{robust_generalizable_rppg_2025}.
FreqPhys further highlights physiological spectral priors as frequency-aware cues for robust rPPG \cite{freqphys2026}.
Together, these studies suggest that temporal waveform constraints and spectral guidance provide complementary supervision for rPPG learning.
However, these studies mainly focus on standard benchmarks, cross-dataset generalization, or general frequency-aware robustness; the appropriate temporal-frequency loss balance under controlled illumination variation remains insufficiently examined.

The contributions of this work are threefold.
First, we formulate rPPG-based heart-rate estimation as a camera-based physiological perception problem for human-facing robotic systems under illumination variation.
Second, our estimator combines PRNet-based facial alignment, clip-level illumination augmentation, and Residual Temporal Standardization Module (RTSM) to reduce illumination-induced temporal feature-statistic shifts.
Third, we study tuning of temporal-frequency supervision on a subject-independent, multi-illumination protocol as a first step toward robust physiological sensing.
We compare time-only, frequency-only, and multiple weighted hybrid settings in a static all-level mix protocol to identify a practical balance between temporal waveform consistency and spectral HR guidance under controlled illumination variation while keeping motion disturbance minimal.

\section{Related Work}
Physiological sensing has been studied as a way to support human-robot interaction and collaboration by estimating human state beyond externally visible behavior.
Prior work on physiological computing in human-robot collaboration has highlighted signals such as heart activity, skin conductance, respiration, and brain activity as cues for workload, stress, affect, and safety \cite{savur2023physiological}.
Non-contact pulse measurement has also been demonstrated on a mobile service robot, showing the relevance of camera-based heart-rate sensing for robotic platforms \cite{stricker2014mobile}.

Early rPPG studies first demonstrated that pulse-related color variations can be measured remotely from ordinary videos under ambient light \cite{verkruysse2008remote}.
Subsequent handcrafted methods improved pulse extraction by separating RGB traces with blind source separation \cite{poh2010noncontact} or by projecting color traces in chrominance and skin-reflection spaces, as in CHROM and POS \cite{dehaan2013robust,wang2017algorithmic}.
These methods established the optical and signal-processing basis of rPPG, but their reliance on color-projection and skin-reflection assumptions makes illumination variation a challenge.
Wang (2020) \cite{Wang2020noncontact} established the dataset protocol used in our lab with controlled illumination levels, static and dynamic tasks, and synchronized ECG/PPG references.
This protocol provides the experimental basis for studying cross-video illumination variation under controlled acquisition conditions.

Recent deep rPPG methods increasingly learn spatiotemporal representations directly from video.
PhysFormer introduced temporal-difference multi-head self-attention (TD-MHSA) and spatio-temporal feed-forward (ST-FF) blocks for facial video-based physiological measurement, making it a suitable predictor backbone for subtle pulse-related spatio-temporal variations \cite{yu2022physformer}.
Illumination-focused assessments further show that lighting variation can strongly affect both traditional and deep rPPG methods \cite{yang2021assessment}.
Hybrid temporal-frequency supervision has also become common in AutoHR, LSTC-rPPG, recent robust rPPG frameworks, and FreqPhys \cite{yu2020autohr,lee2024lstcrppg,robust_generalizable_rppg_2025,freqphys2026}.
These studies motivate the temporal-frequency hybrid loss used in this work, but they do not directly determine the best temporal-frequency balance under controlled illumination variation.

Face alignment is another important factor because unstable facial regions directly affect the quality of the recovered rPPG signal.
PRNet estimates dense 3D face geometry through UV position map regression \cite{feng2018PRNet}.
Wang et al. \cite{wang2025comparative} reported better HR-estimation performance after adopting PRNet-based 3D alignment in dynamic rPPG conditions.
Therefore, this paper uses PRNet as the preprocessing backbone and focuses the main contribution on illumination-aware training and loss design.

Inspired by temporal normalization for robust rPPG \cite{wang2024tnmodule}, we design the Residual Temporal Standardization Module.
Cross-video illumination changes can induce distribution shifts in intermediate temporal feature statistics, which may obscure weak pulse-related patterns.
RTSM makes temporal standardization a residual correction rather than a replacement of the original feature stream, allowing our estimator to suppress illumination-related temporal bias while preserving subtle physiological patterns.

\section{Methodology}
\subsection{Dataset Protocol}
The experiments use the dataset collected in our lab, which follows the protocol described by Wang (2020) \cite{Wang2020noncontact}.
The dataset contains 75 subjects and covers static, speaking, head-rotation, and bicycle tasks under controlled illumination.
Following an 80\% training and 20\% validation protocol, the split contains 60 training subjects and 15 validation subjects.
The current study focuses on cross-video illumination variation; therefore, the completed experiments use static videos from illumination levels 1, 3, and 5 mixed for training and validation.

\begin{table}[t]
\caption{Illumination levels.
The current static mix experiment uses levels 1, 3, and 5 to cover low, medium, and high illumination conditions.
The lux meter is put close to the face of the subject and face toward camera1, camera2, or camera4 successively.}
\label{tab:illumination_levels}
\centering
\footnotesize
\begin{tabular}{cc}
\hline
Illum. & Illuminance at different angle\\
level & (front, forward right, right)\\
\hline
1 & $(40, 20, 10) \pm 50$ LUX\\
3 & $(200, 150, 140) \pm 50$ LUX\\
5 & $(700, 500, 300) \pm 50$ LUX\\
\hline
\end{tabular}
\end{table}

\begin{figure}[t]
\centering
\includegraphics[width=\columnwidth]{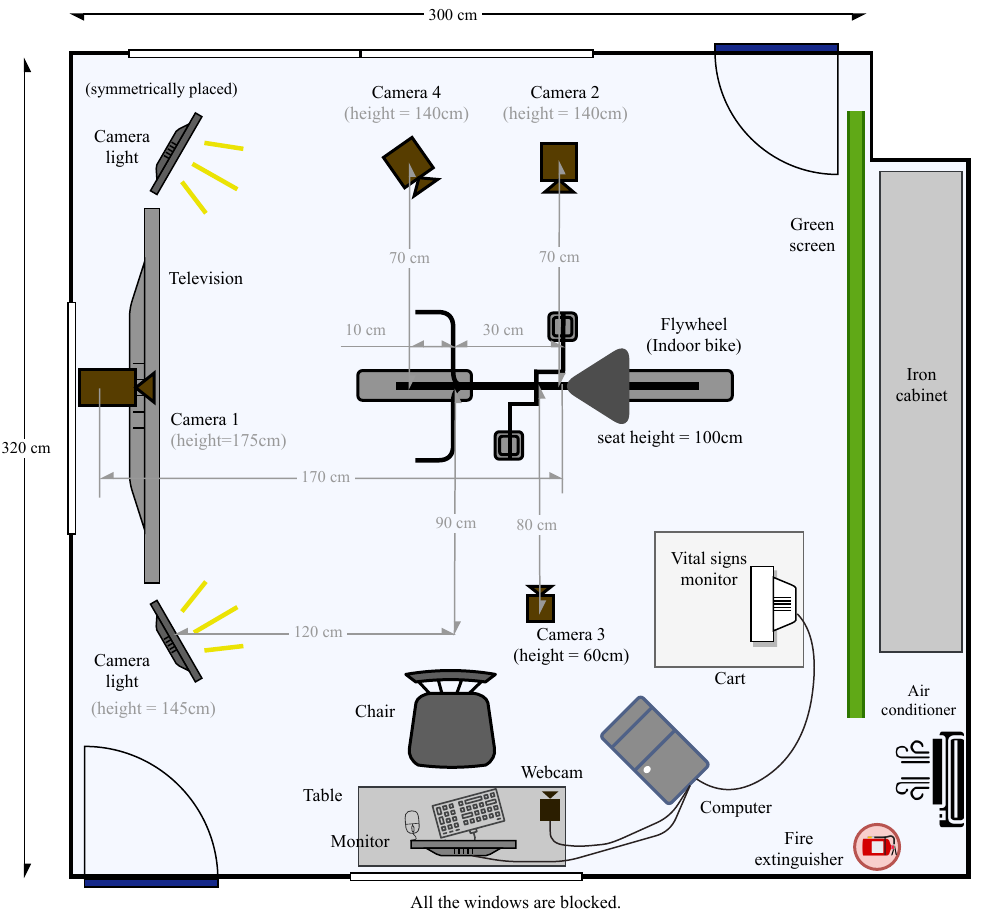}
\caption{Experimental setup for the non-contact physiological signal measurement protocol.
Participants sit on the bike facing the camera mounted above the TV under controlled illumination. Reproduced from Wang (2020) \cite{Wang2020noncontact}.}
\label{fig:experiment_setup}
\end{figure}

In the illumination setting, levels 1 use two camera lights only, while levels 3 and 5 use two camera lights together with the fluorescent light calibrated to produce the illumination in Table \ref{tab:illumination_levels}.
The current study follows the original protocol and uses illumination levels 1, 3, and 5 to represent low, medium, and high illumination conditions in the static all-level mix experiment.
In the static tests, participants sit still on the bike and face the camera above the TV for 2 minutes at each of the three illumination levels, allowing the protocol to isolate the effect of illumination level.
The experimental setup is shown in Fig.~\ref{fig:experiment_setup}.

The current paper focuses on the static all-level mix protocol as a controlled proxy for a common human-facing robot interaction condition in which a user is seated or standing in front of a service or assistive robot camera, while indoor illumination differs across spaces or sessions.
We focus on static videos to isolate illumination variation before studying the additional motion and exercise factors present in speaking, head-rotation, and bicycle tasks, which we also recorded.
These dynamic conditions are left for future analysis of illumination robustness under motion-related and exercise-related factors.

\subsection{Preprocessing}
Raw videos are first processed by a PRNet-based 3D face alignment pipeline.
PRNet predicts dense facial geometry and provides semantically aligned facial regions through UV position map regression \cite{feng2018PRNet}.
In our preprocessing stage, the aligned facial colormap video is exported at $512 \times 512$ resolution.
During model training, the data loader reads the preprocessed videos and resizes each frame to $128 \times 128$ to match the input configuration of our estimator.

Each training clip contains 500 frames, corresponding to 10 seconds at 50 fps.
Pixel values are normalized before being passed into the network, and the ground-truth PPG sequence is normalized per clip.
PRNet is an offline alignment and colormap export stage, while resizing, normalization, and augmentation are performed by the training data loader.

\subsection{Illumination Augmentation}
\label{sec:illumination_aug}
To simulate cross-video illumination variation, the data loader applies clip-level brightness and contrast perturbation.
Given an RGB frame $I_t$, the augmented frame is computed as
\begin{equation}
I'_t = \mathrm{clip}(\alpha_c I_t + \beta_b),
\end{equation}
where the contrast factor is sampled as $\alpha_c \in [0.6, 1.4]$ and the brightness offset is sampled as $\beta_b \in [-40, 40]$.
The same $\alpha_c$ and $\beta_b$ are applied to all frames within a clip.

This clip-level design is used because frame-level random brightness changes can create artificial temporal flicker.
Such flicker would contaminate the temporal signal and make the model learn augmentation artifacts rather than physiological variation.
Clip-level augmentation instead changes the global video appearance while preserving within-clip temporal continuity.

\subsection{Spatial-Temporal Transformer Architecture}
\begin{figure}[t]
\centering
\includegraphics[width=\columnwidth,trim=170 110 150 70,clip]{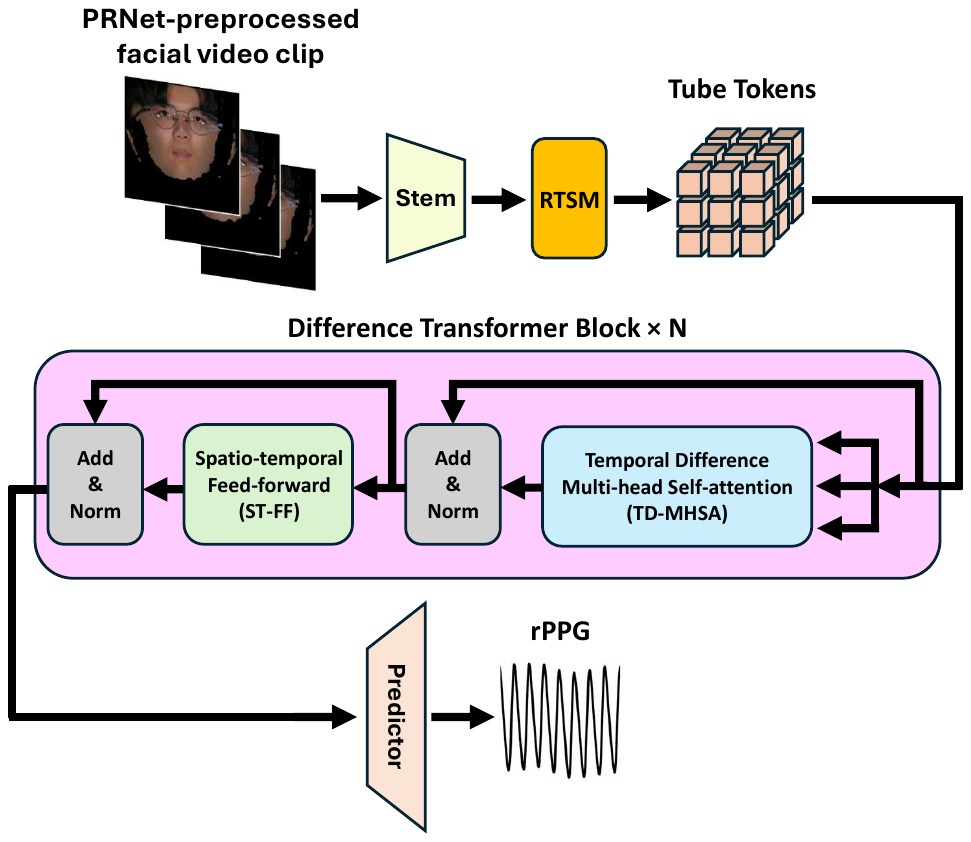}
\caption{Overall architecture of our estimator.
The input is a PRNet-preprocessed facial video clip.
RTSM is the key module in this work; it is inserted after the convolutional stem and before tube-token embedding to reduce brightness-induced temporal feature-statistic shifts.
The PhysFormer-style temporal-difference Transformer block is repeated $N$ times; in our implementation, $N=12$ and the blocks are grouped into three transformer stages.
The final predictor outputs the rPPG waveform.
Clip-level illumination augmentation is not shown because it is used only for data augmentation during training.}
\label{fig:module_architecture}
\end{figure}

We adopt a PhysFormer-style temporal-difference Transformer predictor as the backbone because TD-MHSA and ST-FF blocks are effective at modeling subtle pulse-related spatio-temporal variations from facial videos \cite{yu2022physformer}.
These PhysFormer-style blocks are used as the predictor backbone, while RTSM provides the illumination-aware feature correction studied in this work.
Our estimator takes in a PRNet-preprocessed facial video clip with input shape $[B,3,500,128,128]$, where $B$ is batch size.
A 3D convolutional stem first extracts shallow spatiotemporal features.
RTSM is inserted after the convolutional stem and before tube-token embedding to stabilize intermediate temporal feature statistics, and is detailed in the next subsection.
After RTSM, the features are converted into tube tokens and processed by PhysFormer-style Transformer stages, followed by temporal upsampling, spatial averaging, and a final one-dimensional convolution to predict the rPPG waveform.
Fig.~\ref{fig:module_architecture} summarizes the overall architecture of our estimator.

\subsection{Residual Temporal Standardization Module}
\label{sec:rtsm}
\begin{figure}[!tbhp]
\centering
\includegraphics[width=0.85\columnwidth]{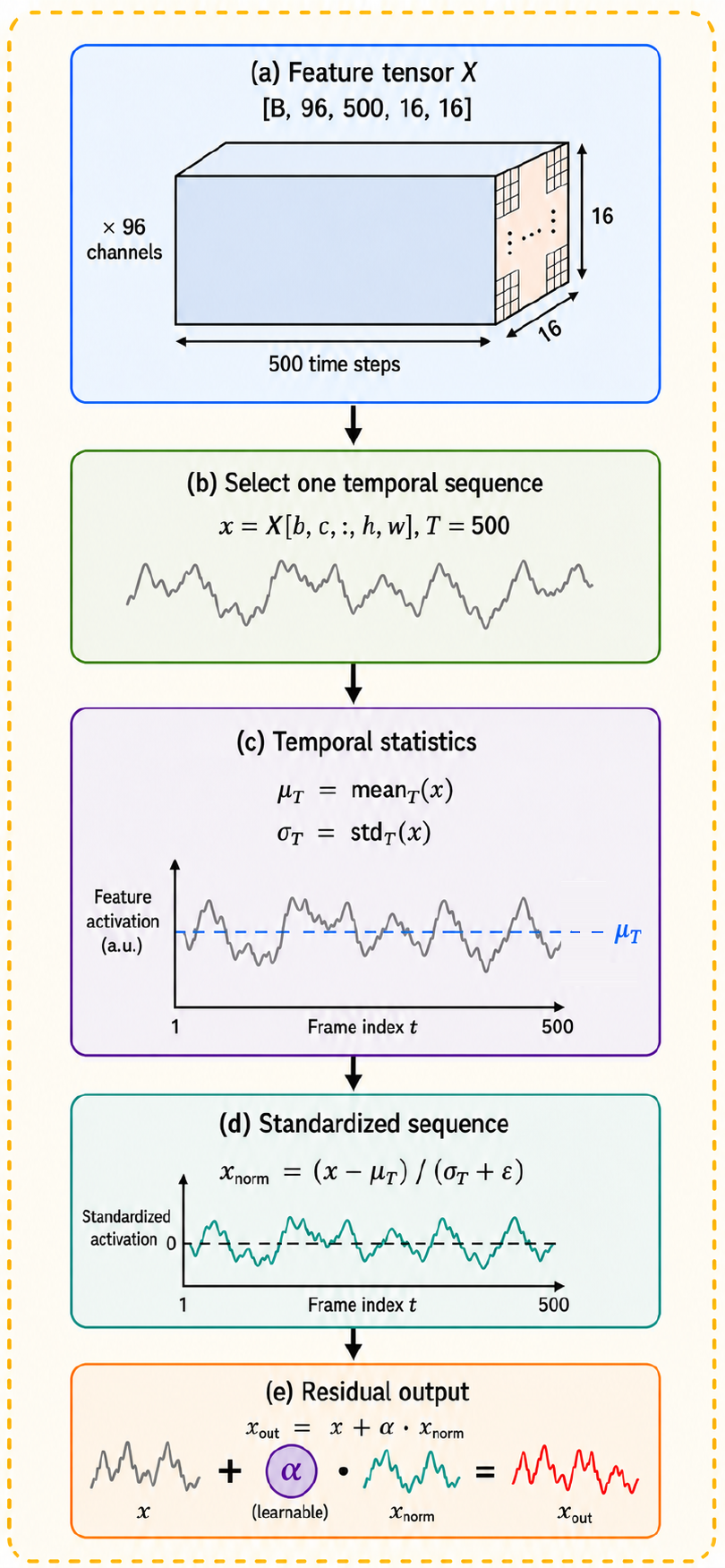}
\caption{Visualization of the Residual Temporal Standardization Module.
(a) It operates on the stem feature tensor $X$ with shape $[B,96,500,16,16]$.
(b) One temporal feature sequence $x=X[b,c,:,h,w]$ is selected for visualization.
(c) The temporal mean $\mu_T$ and standard deviation $\sigma_T$ are computed over $T=500$ frames for each channel and spatial location rather than globally over all channels or spatial positions.
(d) The selected sequence is standardized as $x_{\mathrm{norm}}=(x-\mu_T)/(\sigma_T+\epsilon)$.
(e) A learnable residual scale $\alpha$, optimized by backpropagation, modulates $x_{\mathrm{norm}}$ to produce $x_{\mathrm{out}}=x+\alpha x_{\mathrm{norm}}$.}
\label{fig:rtsm_visualization}
\end{figure}

The Residual Temporal Standardization Module (RTSM) is designed to reduce brightness-induced distribution shifts in intermediate temporal feature statistics before the PhysFormer-style Transformer predictor.
Let $X$ denote the feature tensor after the convolutional stem.
In our implementation, $X \in \mathbb{R}^{B \times 96 \times 500 \times 16 \times 16}$.
RTSM computes temporal statistics along the temporal dimension for each channel and spatial location:
\begin{equation}
X_{\mathrm{norm}} =
\frac{X-\mu_T(X)}{\sigma_T(X)+\epsilon},
\end{equation}
\begin{equation}
X_{\mathrm{out}} = X + \alpha X_{\mathrm{norm}}. \label{eq:rtsm}
\end{equation}
Here, $\mu_T$ and $\sigma_T$ are computed along the temporal dimension $T=500$ for each channel and spatial location, $\epsilon$ is set to $10^{-5}$ for numerical stability, and $\alpha$ is a learnable residual scaling parameter initialized to 0.01.
Because $\alpha$ is learnable and unconstrained, RTSM can adaptively amplify or suppress the standardized temporal component.
The residual design stabilizes temporal statistics without erasing the original feature representation.
Fig.~\ref{fig:rtsm_visualization} visualizes the operation for one temporal feature sequence $x=X[b,c,:,h,w]$ selected from the full tensor; lowercase $x$ is used only for this realization, whereas uppercase $X$ denotes the full feature tensor in the equations.

\subsection{Hybrid Temporal-Frequency Loss}
\label{sec:hybrid_loss}
The training objective is a hybrid temporal-frequency loss:
\begin{equation}
L_{\mathrm{total}} = L_{\mathrm{time}} + \beta L_{\mathrm{freq}},
\label{eq:total_loss}
\end{equation}
where $L_{\mathrm{time}}$ is a Soft-Shifted Pearson loss and $L_{\mathrm{freq}}$ is a spectral Kullback-Leibler (KL) divergence loss.
The weight $\beta$ controls the contribution of frequency-domain supervision.

For the time-domain term, let $\hat{y}$ be the predicted rPPG waveform and $y$ be the ground-truth waveform.
To tolerate small temporal offsets between prediction and ground truth, we compute the Pearson correlation over a shift set $\mathcal{S}=\{-S,\ldots,S\}$.
In our implementation, $S=3$ frames, corresponding to a maximum temporal tolerance of 0.06 s at the 50 fps sampling rate.
The shift set therefore allows the loss to compensate for minor phase offsets caused by face alignment, clip sampling, or temporal filtering, while keeping the maximum shift much shorter than a cardiac cycle.
For each sample $b$ in the batch and shift $s$,
\begin{equation}
\rho_{b,s} =
\frac{\sum_t(\hat{y}_{b,t+s}-\bar{\hat{y}}_{b,s})(y_{b,t}-\bar{y}_b)}
{\sqrt{
\left(\sum_t(\hat{y}_{b,t+s}-\bar{\hat{y}}_{b,s})^2\right)
\,
\left(\sum_t(y_{b,t}-\bar{y}_b)^2\right)
+\epsilon}}.
\end{equation}
The numerical-stability constant is set to $\epsilon=10^{-8}$.
The Soft-Shifted Pearson loss then uses a softmax-weighted average of the shifted Pearson correlations:
\begin{equation}
w_{b,s} =
\frac{\exp(\rho_{b,s}/\tau)}
{\sum_{q\in\mathcal{S}}\exp(\rho_{b,q}/\tau)},
\end{equation}
\begin{equation}
L_{\mathrm{time}}
=
1-\frac{1}{B}\sum_{b=1}^{B}
\sum_{s\in\mathcal{S}} w_{b,s}\rho_{b,s}.
\end{equation}
Here, $b$ indexes a sample in the batch, $B$ is the batch size used in each forward pass, and $q$ indexes the candidate shifts in $\mathcal{S}$.
In our experiments, $B=2$.
The temperature parameter is set to $\tau=0.1$; smaller values make the weighting concentrate more strongly on the shift with the highest Pearson correlation.

In this work, each waveform denotes a 500-sample temporal sequence at 50 fps, either the predicted rPPG signal or the ground-truth PPG signal for the same video clip.
For the frequency-domain term, the predicted and ground-truth waveforms are transformed into one-sided power spectra.
Before the Fourier transform, each waveform is mean-centered and multiplied by a length-500 Hann window to reduce spectral leakage caused by finite-length clip boundaries.
A single full-clip FFT is applied to each 500-frame waveform; therefore, no sliding-window overlap is used.
For HR metric computation, the dominant frequency is searched within 40--180 bpm, corresponding to 0.67--3.00 Hz.
For the frequency-domain loss, the KL target distribution is centered on the dominant spectral peak of the ground-truth waveform.
The spectra are normalized into distributions $P_{\hat{y}}$ and $P_y$.
The spectral KL loss is
\begin{equation}
L_{\mathrm{freq}} =
\sum_k P_y(k)\log\frac{P_y(k)+\epsilon}{P_{\hat{y}}(k)+\epsilon}.
\end{equation}
The same numerical-stability constant, $\epsilon=10^{-8}$, is used in the KL computation.
This term encourages the predicted waveform spectrum to align with the HR-related spectral distribution derived from the ground-truth signal.

\section{Results}
We evaluate our estimator on static videos from three illumination levels.
All HR performance metrics reported in this section are computed on the full validation set.

To determine the best temporal-frequency balance, we evaluate the frequency weight $\beta$ in \eqref{eq:total_loss} at six values from time-only ($\beta=0$) to frequency-only, with three independent training runs per setting.
$\beta=5$ yields the lowest HR MAE and RMSE and the highest correlation in both best-run and mean results (Table~\ref{tab:static_mix}).
Relative to time-only supervision, $\beta=5$ reduces HR MAE by 0.92~bpm and HR RMSE by 2.47~bpm; relative to frequency-only supervision, the reductions are 1.46~bpm and 3.03~bpm, respectively.
Neither temporal nor spectral supervision alone is sufficient; combining both terms consistently improves performance.
The weaker result at $\beta=6$ indicates that further increasing the frequency weight is not beneficial, suggesting that the model requires strong but balanced spectral guidance under mixed illumination.

\begin{table}[t]
\caption{Static All-Level Mix Performance Analysis ($\downarrow$ lower is better, $\uparrow$ higher is better)}
\label{tab:static_mix}
\centering
\scriptsize
\begin{tabular*}{\columnwidth}{@{\extracolsep{\fill}}lccccccc@{}}
\hline
\multirow{2}{*}{Setting} & \multirow{2}{*}{$n$} & \multicolumn{2}{@{}c@{}}{HR MAE$\downarrow$} & \multicolumn{2}{@{}c@{}}{HR RMSE$\downarrow$} & \multicolumn{2}{@{}c@{}}{HR $r\uparrow$} \\
 & & min & mean & min & mean & max & mean \\
\hline
Time-only & 3 & 1.71 & 2.32 & 4.87 & 5.95 & 0.926 & 0.891 \\
$\beta=1$ & 3 & 1.37 & 1.43 & 3.50 & 4.05 & 0.945 & 0.949 \\
$\beta=2$ & 3 & 1.33 & 1.43 & 4.24 & 4.31 & 0.946 & 0.943 \\
$\beta=3$ & 3 & 1.21 & 1.32 & 3.43 & 3.73 & 0.964 & 0.957 \\
$\beta=4$ & 3 & 1.25 & 1.42 & 3.74 & 4.06 & 0.958 & 0.951 \\
$\beta=5$ & 3 & $\mathbf{0.79}$ & $\mathbf{0.99}$ & $\mathbf{2.40}$ & $\mathbf{3.09}$ & $\mathbf{0.982}$ & $\mathbf{0.970}$ \\
$\beta=6$ & 3 & 1.25 & 1.33 & 3.39 & 3.63 & 0.965 & 0.959 \\
Freq-only & 3 & 2.25 & 2.61 & 5.43 & 6.60 & 0.915 & 0.883 \\
\hline
\end{tabular*}
\end{table}

\begin{table}[t]
\caption{Comparison with the PhysFormer Baseline and Component Ablation under $\beta=5$ Hybrid Supervision}
\label{tab:combined_ablation}
\centering
\scriptsize
\begin{tabular*}{\columnwidth}{@{}l@{\extracolsep{\fill}}c@{\hspace{0.55em}}c@{\hspace{0.55em}}c@{\hspace{1.6em}}c@{\hspace{0.75em}}c@{\hspace{0.75em}}c@{}}
\hline
\multirow{2}{*}{Method} & Hybrid & \multirow{2}{*}{Aug.} & \multirow{2}{*}{RTSM} & \multirow{2}{*}{MAE$\downarrow$} & \multirow{2}{*}{RMSE$\downarrow$} & \multirow{2}{*}{$r\uparrow$} \\
 & Loss & & & & & \\
\hline
PhysFormer \cite{yu2022physformer} & -- & -- & -- & 12.33 & 15.51 & 0.088 \\
Hybrid loss only & $\checkmark$ & -- & -- & 3.54 & 8.99 & 0.778 \\
+ Augmentation & $\checkmark$ & $\checkmark$ & -- & 2.33 & 5.79 & 0.912 \\
+ RTSM & $\checkmark$ & -- & $\checkmark$ & 1.25 & 3.08 & 0.970 \\
Full estimator & $\checkmark$ & $\checkmark$ & $\checkmark$ & $\mathbf{0.79}$ & $\mathbf{2.40}$ & $\mathbf{0.982}$ \\
\hline
\end{tabular*}
\end{table}

We fix $\beta=5$ for the component ablation study (Table~\ref{tab:combined_ablation}).
The hybrid-loss-only variant reduces HR MAE from 12.33~bpm to 3.54~bpm compared with the PhysFormer baseline, a 71.3\% relative reduction.
Adding RTSM alone further reduces HR MAE to 1.25~bpm and raises HR correlation to 0.970; adding illumination augmentation alone reduces HR MAE to 2.33~bpm.
Combining both components yields 0.79~bpm HR MAE, 2.40~bpm HR RMSE, and 0.982 HR correlation, corresponding to 93.6\% and 84.5\% reductions in HR MAE and RMSE relative to the PhysFormer baseline.
The hybrid loss accounts for the largest single improvement, while RTSM and illumination augmentation provide complementary gains under varying lighting conditions.

\begin{figure}[!tbhp]
\centering
\includegraphics[width=0.82\columnwidth,trim=0 4 0 4,clip]{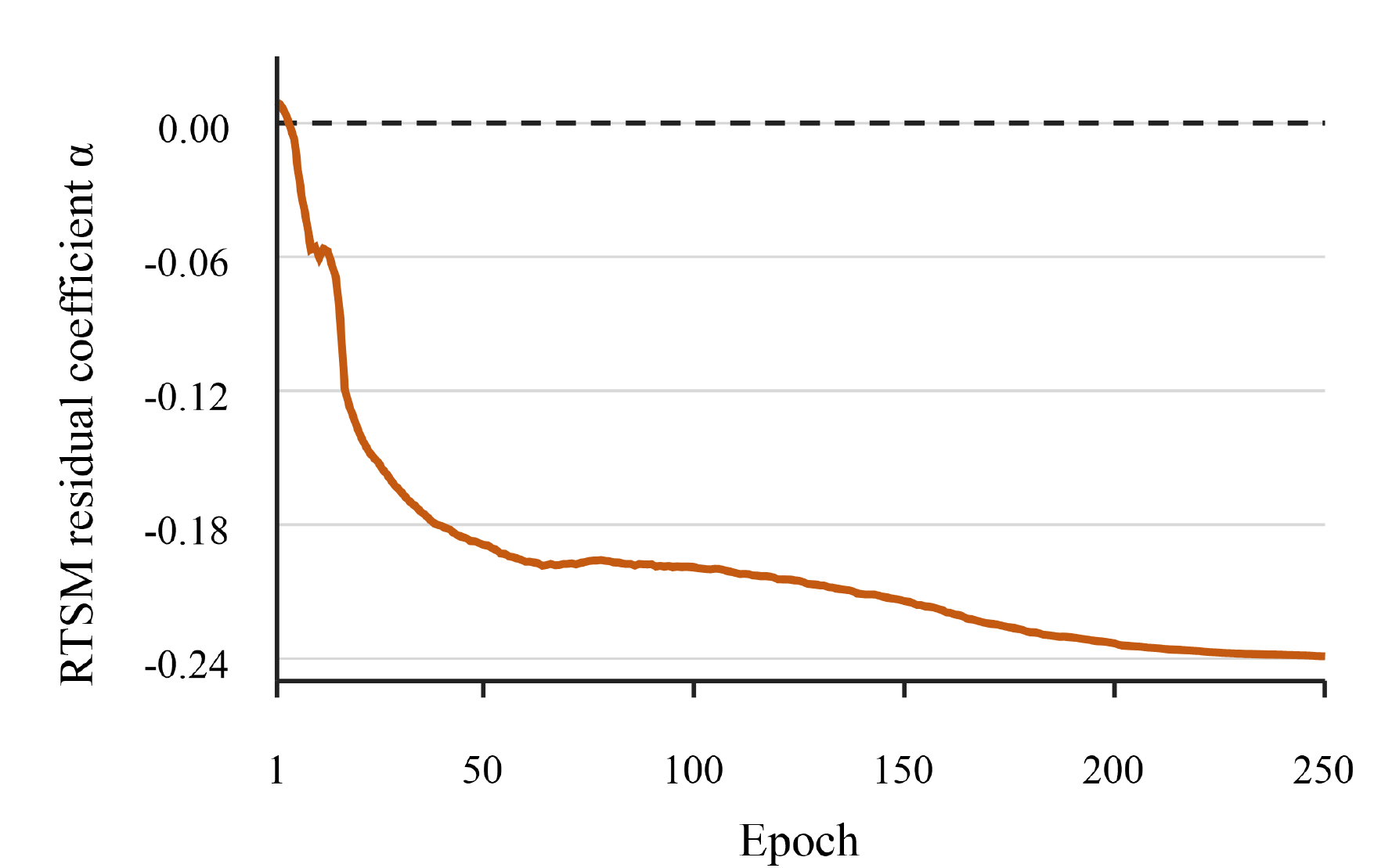}
\caption{Learned RTSM residual coefficient $\alpha$.}
\label{fig:rtsm_alpha_distribution}
\end{figure}

The learnable residual coefficient $\alpha$ in (\ref{eq:rtsm}) moved toward negative values during optimization (Fig.~\ref{fig:rtsm_alpha_distribution}).
Since $X_{\mathrm{norm}}$ preserves the peaks and valleys of $X$, a negative $\alpha$ subtracts a scaled copy of the same temporal pattern, reducing the amplitude of non-physiological variation while retaining the pulse-related waveform shape.

\section{Discussion}
The static all-level mix results show that illumination robustness depends on both data-side and objective-side design.
PRNet preprocessing provides stable aligned facial inputs, which helps reduce spatial inconsistency before learning.
Clip-level illumination augmentation expands the apparent brightness and contrast distribution while preserving temporal continuity.
RTSM then stabilizes temporal feature statistics inside our estimator, following the robustness motivation of temporal normalization \cite{wang2024tnmodule} while adapting the structure to the current end-to-end framework.

The negative $\alpha$ values indicate that RTSM learns to use the standardized temporal component as a suppressive residual correction, reducing non-physiological temporal variation while preserving the original feature stream through the residual pathway.
Prior work has shown that temporal waveform constraints and spectral HR guidance are complementary, but the balance under controlled illumination variation has not been examined.
In this static all-level mix protocol, frequency guidance needs to be strong enough to stabilize HR estimation when the same subject-independent validation split contains multiple illumination levels.
At the same time, the weaker result at $\beta=6$ shows that increasing the frequency weight further is not beneficial.
Among the tested beta settings, $\beta=5$ achieves the best HR MAE, RMSE, and correlation, showing that the weight between temporal and frequency loss benefits from optimization.

This study does not yet evaluate moving robot platforms, online adaptation, or closed-loop HRI behavior.
Instead, the current static all-level mix protocol isolates illumination variation as a controlled first step toward robot-facing physiological sensing.
Future work will study how illumination, motion, and physiological-state changes interact in speaking, head-rotation, bicycle-exercise, and robot-mounted-camera conditions.

\section{Conclusion}
This paper demonstrates an estimator for camera-based physiological perception through rPPG-based heart-rate estimation under controlled illumination variation.
Our estimator integrates PRNet-based 3D face alignment, clip-level illumination augmentation, RTSM, and hybrid temporal-frequency supervision.
On our dataset, the static all-level mix experiment shows that $\beta=5$ gives the strongest performance among the tested weights, achieving 0.79 bpm HR MAE, 2.40 bpm HR RMSE, and 0.982 HR correlation.
Compared with the PhysFormer baseline on the same protocol, our estimator reduces HR MAE by 93.6\% and HR RMSE by 84.5\%.
These results support the value of combining objective-level hybrid supervision, feature-level residual temporal standardization, and data-side illumination augmentation for robust rPPG learning.
These findings support illumination-robust rPPG as a promising component for unobtrusive physiological perception in future human-facing robotic systems.

\section*{Acknowledgments}
This work was supported by the Taiwan National Science and Technology Council (NSTC) under Grant 111-2221-E-006-186 and Grant 114-2221-E-006-089.

\section*{Declaration on Generative AI}
During the preparation of this work, the authors used Claude (Anthropic) in order to: Grammar and spelling check, code generation for figure plotting scripts, and LaTeX formatting assistance.
After using this tool, the authors reviewed and edited the content as needed and take full responsibility for the publication's content.

\bibliographystyle{unsrtnat}
\bibliography{references}

@article{savur2023physiological,
  author = {Savur, Celal and Sahin, Ferat},
  journal = {Machines},
  note = {doi: 10.3390/machines11050536},
  number = {5},
  pages = {536},
  title = {Survey on Physiological Computing in Human-Robot Collaboration},
  volume = {11},
  year = {2023},
}

@article{verkruysse2008remote,
  author = {Verkruysse, Wim and Svaasand, Lars O and Nelson, J Stuart},
  publisher = {Optical Society of America},
  journal = {Optics express},
  note = {doi: 10.1364/OE.16.021434},
  number = {26},
  pages = {21434--21445},
  title = {Remote plethysmographic imaging using ambient light.},
  volume = {16},
  year = {2008},
}

@inproceedings{stricker2014mobile,
  author = {Stricker, Ronny and Mueller, Steffen and Gross, Horst-Michael},
  booktitle = {Proceedings of the 23rd IEEE International Symposium on Robot and Human Interactive Communication},
  note = {doi: 10.1109/ROMAN.2014.6926392},
  pages = {1056--1062},
  title = {Non-Contact Video-Based Pulse Rate Measurement on a Mobile Service Robot},
  year = {2014},
}

@article{yu2020autohr,
  author = {Yu, Zitong and Li, Xiaobai and Niu, Xuesong and Shi, Jingang and Zhao, Guoying},
  publisher = {IEEE},
  journal = {IEEE Signal Processing Letters},
  note = {doi: 10.1109/LSP.2020.3007086},
  pages = {1245--1249},
  title = {{AutoHR}: A strong end-to-end baseline for remote heart rate measurement with neural searching},
  volume = {27},
  year = {2020},
}

@inproceedings{yu2022physformer,
  author = {Yu, Zitong and Shen, Yuming and Shi, Jingang and Zhao, Hengshuang and Torr, Philip HS and Zhao, Guoying},
  booktitle = {Proceedings of the IEEE/CVF Conference on Computer Vision and Pattern Recognition},
  note = {doi: 10.1109/CVPR52688.2022.00415},
  pages = {4186--4196},
  title = {{PhysFormer}: facial video-based physiological measurement with temporal difference transformer},
  year = {2022},
}

@inproceedings{lee2024lstcrppg,
  author = {Lee, Jun Seong and Hwang, Gyutae and Ryu, Moonwook and Lee, Sang Jun},
  booktitle = {Proceedings of the IEEE/CVF Conference on Computer Vision and Pattern Recognition Workshops},
  note = {doi: 10.1109/CVPRW59228.2023.00640},
  title = {{LSTC-rPPG}: Long Short-Term Convolutional Network for Remote Photoplethysmography},
  year = {2024},
}

@article{robust_generalizable_rppg_2025,
  author = {Cen, Kang and Fu, Chang-Hong and Hong, Hong},
  journal = {arXiv preprint},
  note = {doi: 10.48550/arXiv.2507.07795},
  title = {Robust and Generalizable Heart Rate Estimation via Deep Learning for Remote Photoplethysmography in Complex Scenarios},
  year = {2025},
}

@article{freqphys2026,
  author = {Qian, Wei and Guo, Dan and Zhou, Jinxing and Zou, Bochao and Yu, Zitong and Wang, Meng},
  journal = {arXiv preprint},
  note = {doi: 10.48550/arXiv.2604.00534},
  title = {{FreqPhys}: Repurposing Implicit Physiological Frequency Prior for Robust Remote Photoplethysmography},
  year = {2026},
}

@article{poh2010noncontact,
  author = {Poh, Ming-Zher and McDuff, Daniel J. and Picard, Rosalind W.},
  journal = {Optics Express},
  note = {All Open Access, Gold Open Access; doi: 10.1364/OE.18.010762},
  number = {10},
  pages = {10762--10774},
  title = {Non-contact, automated cardiac pulse measurements using video imaging and blind source separation},
  type = {Article},
  volume = {18},
  year = {2010},
}

@article{dehaan2013robust,
  author = {de Haan, Gerard and Jeanne, Vincent},
  journal = {IEEE Transactions on Biomedical Engineering},
  note = {doi: 10.1109/TBME.2013.2266196},
  number = {10},
  pages = {2878--2886},
  title = {Robust Pulse Rate From Chrominance-Based rPPG},
  volume = {60},
  year = {2013},
}

@article{wang2017algorithmic,
  author = {Wang, Wenjin and den Brinker, Albertus C. and Stuijk, Sander and de Haan, Gerard},
  journal = {IEEE Transactions on Biomedical Engineering},
  note = {doi: 10.1109/TBME.2016.2609282},
  number = {7},
  pages = {1479--1491},
  title = {Algorithmic Principles of Remote PPG},
  volume = {64},
  year = {2017},
}

@mastersthesis{Wang2020noncontact,
  author = {Wang, Chien-Chih},
  address = {No. 1, Dasyue Rd, East District, Tainan City, 701},
  school = {National Cheng Kung University},
  title = {{Non-contact heart rate measurement based on facial videos}},
  year = {2020},
}

@article{yang2021assessment,
  author = {Yang, Ze and Wang, Haofei and Lu, Feng},
  journal = {IEEE Transactions on Human-Machine Systems},
  note = {doi: 10.1109/THMS.2022.3207755},
  number = {6},
  pages = {1236--1246},
  title = {Assessment of Deep Learning-Based Heart Rate Estimation Using Remote Photoplethysmography Under Different Illuminations},
  volume = {52},
  year = {2022},
}

@inproceedings{feng2018PRNet,
  author = {Feng, Yao and Wu, Fan and Shao, Xiaohu and Wang, Yanfeng and Zhou, Xi},
  booktitle = {Proceedings of the European conference on computer vision (ECCV)},
  note = {doi: 10.1007/978-3-030-01264-9\_32},
  pages = {534--551},
  title = {Joint 3D Face Reconstruction and Dense Alignment with Position Map Regression Network},
  year = {2018},
}

@mastersthesis{wang2025comparative,
  author = {Wang, Yu-Chiao},
  address = {Tainan, Taiwan},
  month = {6},
  school = {National Cheng Kung University},
  title = {Comparative Analysis of Non-End-to-End and End-to-End Deep Learning Models with 2D and 3D Face Alignment for Remote Heart Rate Estimation},
  year = {2025},
}

@article{wang2024tnmodule,
  author = {Wang, Kegang and Tang, Jiankai and Wei, Yantao and Liu, Mingxuan and Liu, Xin and Wang, Yuntao},
  journal = {arXiv preprint},
  note = {doi: 10.48550/arXiv.2411.15283},
  title = {A Plug-and-Play Temporal Normalization Module for Robust Remote Photoplethysmography},
  year = {2024},
}

\end{document}